\pdfoutput=1
\documentclass[mlmain,onecolumn]{jmlr}
\jmlrpmlr %

\usepackage{booktabs}
\providecommand{\doi}[1]{%
  doi: \begingroup\urlstyle{rm}\def\UrlBreaks{\do\/}%
  \href{https://doi.org/#1}{\nolinkurl{#1}}\endgroup}

\graphicspath{{figures/}}

\jmlrvolume{}
\firstpageno{1}
\editors{}
\jmlryear{2026}
\jmlrworkshop{Symmetry and Geometry in Neural Representations}
\title[ConceptBasis]{ConceptBasis}
\author{\Name{Xinyun Zhou} \Email{markjouh@icloud.com}\\
  \addr University of Wisconsin--Madison
  \AND
  \Name{Walter Gerych} \Email{wgerych@wpi.edu}\\
  \addr Worcester Polytechnic Institute}
\hypersetup{pdfauthor={Xinyun Zhou, Walter Gerych},pdftitle={ConceptBasis}}

\begin{document}
\maketitle

\begin{abstract}
The linear representation hypothesis proposes that semantic concepts are encoded as directions in a model's embedding space. Directions estimated with linear probes or differences between group means already support useful concept readout, steering, and debiasing. However, directions that work individually need not form a separable, additive coordinate system. Such a coordinate system would make output embeddings natively interpretable and enable more precise test-time interventions. Group-mean-derived directions can be confounded by co-occurring concepts, while directions estimated after controlling for those co-occurrences can still overlap geometrically. We introduce ConceptBasis, which constructs a fixed concept dictionary and dense label matrix, uses joint ridge regression to estimate partial concept effects, and trains lightweight image and text adapters to reduce the remaining overlap while retaining contrastive retrieval. We evaluate direction overlap and class-disjoint compositional retrieval on the THINGS object dataset. Compared with orthogonality training on group-mean-derived directions, ConceptBasis reduces root-mean-square (RMS) overlap from .124 to .057 and raises ten-attribute compositional Recall@5 from .620 to .728, while bidirectional image--text Recall@5 remains virtually unchanged.
\end{abstract}

\section{Introduction}

Many empirical results are consistent with semantic information occupying approximately linear directions in representation space, an idea formalized by the linear representation hypothesis \citep{park2024linear}. Early word embeddings showed that semantic relationships can be expressed through vector arithmetic, as in the famous example of $\text{king}-\text{man}+\text{woman}\approx\text{queen}$ \citep{mikolov2013linguistic}. Later work showed the practical implications of these ideas. Directions estimated with linear probes support concept-based interpretation \citep{kim2018tcav}; differences between group means can steer model behavior \citep{rimsky2024steering}; and projecting out identified directions can reduce bias in word embeddings \citep{bolukbasi2016debiasing} and vision-language models \citep{chuang2023debiasing}. These results establish concept directions as useful interfaces to learned representations.

We ask whether individually useful directions can also form a separable, additive coordinate system. Such a system would make each coefficient a less-confounded measure of its named concept, enable embedding edits with less collateral damage, and allow more natural composition without double-counting shared components. An image embedding may represent a red, round apple while its directions for \texttt{red}, \texttt{round}, and \texttt{apple} share significant components of one another. Shared components can cause cross-talk when a direction is used individually and repeatedly count the same information when directions are combined. Standard vision-language objectives train image--text alignment \citep{radford2021clip,tschannen2025siglip2}, but do not directly encourage either separability or additivity.

\begin{figure}[t]
  \centering
  \includegraphics[width=0.64\linewidth]{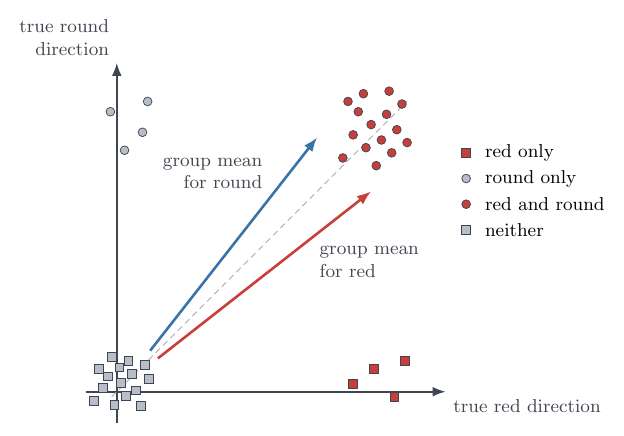}
  \caption{Correlated observations skew group-mean-derived directions toward the diagonal even when the underlying concept directions are orthogonal.}
  \label{fig:group-mean-confounding}
\end{figure}

Two problems stand between individually useful directions and this coordinate system. First, correlations between concepts cause group-mean-derived directions to lean toward each other, even when their true partial effects are orthogonal (Figure~\ref{fig:group-mean-confounding}). Second, controlling for observed co-occurrence addresses this statistical confounding but does not necessarily force the resulting directions to be orthogonal. They may still share components because standard training provides no incentive to separate them.

Linear probes and group means identify directions along which examples with a concept are separated from examples without it. They can use any available signal, including correlated concepts, to make that distinction. This is useful for readout and for interventions intended to suppress group-predictive information, such as debiasing, but it does not isolate the component associated specifically with the concept. We instead seek partial effects: the embedding difference associated with one concept after controlling for the other observed concepts.

In this work, we propose ConceptBasis, which addresses both problems. Controlling for co-occurring concepts requires a label for every concept on every image, so ConceptBasis first builds a concept dictionary by discovering and merging visual tags, then labels each image against it. It estimates partial effects from these labels with joint ridge regression, then trains lightweight image and text adapters with an orthogonality objective to reduce the remaining overlap among the estimated directions. A contrastive objective preserves ordinary image--text retrieval. Because the adapters operate on a frozen vision-language encoder, ConceptBasis reshapes its embedding geometry without retraining the underlying model.

We evaluate ConceptBasis on the THINGS dataset, measuring geometric overlap between partial-effect directions and testing additivity through compositional retrieval. Given a subset of an image's attributes, we sum their concept directions and use the result to retrieve the source image from a gallery of unseen object classes. If the representation is additive, supplying more correct attributes should make the intended image easier to identify. We separately evaluate bidirectional image--text retrieval to test whether the new geometry comes at the cost of the original representation.

Across five seeds, ConceptBasis reduces root-mean-square (RMS) direction overlap from .124 to .057 relative to orthogonality training on group-mean-derived directions. On the held-out test classes, ten-attribute compositional Recall@5 rises from .620 to .728. Bidirectional image--text Recall@5 remains virtually unchanged at .967, compared with .968 for contrastive training alone. These results suggest that individually useful concept directions can be organized into a substantially more separable and additive system without sacrificing ordinary retrieval.

Our contributions are:
\begin{itemize}
  \item We distinguish group-predictive directions, which may exploit correlated concepts, from partial concept effects, and identify statistical confounding and geometric overlap as separate barriers to separability and additivity.
  \item We introduce ConceptBasis, which constructs its concept dictionary and labels, estimates partial concept effects with joint ridge regression, and trains lightweight image and text adapters to reduce their remaining overlap while preserving contrastive alignment.
  \item We introduce a class-disjoint compositional retrieval evaluation on THINGS and show that ConceptBasis improves geometric separability and additive retrieval without sacrificing bidirectional image--text retrieval.
\end{itemize}
Code is available at \href{https://github.com/markjouh/conceptbasis/tree/786074beed8c5d23681731fbddeecc54195f0188}{\texttt{github.com/markjouh/conceptbasis}}.

\section{Related Work}

Semantic vector offsets in word embeddings provided early evidence for linear concept structure \citep{mikolov2013linguistic}, and the linear representation hypothesis now formalizes connections among concept directions, probing, and intervention \citep{park2024linear}. Concept activation vectors introduced human-defined directions for interpreting neural activations \citep{kim2018tcav}, while mean-difference directions provide a particularly simple and effective steering mechanism \citep{rimsky2024steering}. Projection along semantic directions has also enabled post-hoc debiasing in word and vision-language embeddings \citep{bolukbasi2016debiasing,chuang2023debiasing}. These methods seek directions that are useful individually; ConceptBasis instead asks whether a collection of directions can serve as separable, additive coordinates.

Concept whitening decorrelates a layer of a CNN classifier and rotates it so that each axis aligns with a named concept \citep{chen2020conceptwhitening}. Each axis is fit to the mean of its concept's example images, so, like a group-mean direction, it can be confounded by co-occurring concepts: an \texttt{airplane} axis also captures sky. ConceptBasis instead estimates partial effects by joint regression on all concept labels, which controls for co-occurring concepts.

CEDAR rotates the embedding space of a frozen vision-language model such that its embeddings best survive a top-$k$ sparsity bottleneck, then names each axis after its nearest text concept \citep{kubaty2026cedar}. Because a rotation preserves all angles between vectors, it cannot reduce the overlap between concept directions. ConceptBasis instead reshapes the geometry itself, training adapters to reduce this overlap.

SpLiCE decomposes a frozen CLIP embedding into a sparse nonnegative combination of word-named atoms for post-hoc interpretation \citep{bhalla2024splice}. These names need not describe the image: because the atoms are correlated and overcomplete, the decomposition can reconstruct a tiger from \texttt{tiger} while leaving \texttt{animal} at zero. ConceptBasis instead constructs a non-redundant dictionary, merging synonyms and removing near-duplicates, and trains the concept directions toward orthogonality, so that concepts cannot readily substitute for one another in a reconstruction.

\section{Preliminaries}

Let a frozen vision-language encoder map an image $x_i$ and caption $t_i$ to embeddings $f_I(x_i)$ and $f_T(t_i)$; the subscripts $I$ and $T$ denote the image and text modalities throughout. Trainable image and text adapters $g_I$ and $g_T$ transform these into vectors $z_i=g_I(f_I(x_i))$ and $u_i=g_T(f_T(t_i))$ in a shared $p$-dimensional space, and we write $z(x)=g_I(f_I(x))$ for an arbitrary image $x$. Each of the $n$ training images has binary labels $y_i \in \{0,1\}^K$, $i=1,\ldots,n$, over a fixed dictionary of $K$ affirmative visual concepts. In our experiments, $p=K=256$.

For unit concept directions $d_1,\ldots,d_K\in\mathbb{R}^p$, collected as the rows of $D\in\mathbb{R}^{K\times p}$, we define the coefficient of concept $k$ as its projection onto $d_k$, standardized using the mean $\mu_k$ and standard deviation $\sigma_k$ of the training-set projections:
\begin{equation}
  c_k(x)=\frac{z(x)^\top d_k-\mu_k}{\sigma_k}.
  \label{eq:coefficient}
\end{equation}
Because unit directions can still have different projection variances, this keeps the coordinates on comparable scales for additive composition.

A coordinate is readable when $c_k(x)$ ranks held-out examples by whether concept $k$ is present. Potential geometric interference between the directions is summarized by off-diagonal RMS overlap,
\begin{equation}
  \operatorname{overlap}(D)=
  \sqrt{\frac{1}{K(K-1)}\sum_{j\neq k}(d_j^\top d_k)^2}.
  \label{eq:overlap}
\end{equation}
For a concept set $S$, additive retrieval scores a gallery image by $r(x\mid S)=\sum_{k\in S}c_k(x)$. This asks whether concepts can be combined while maintaining fidelity, rather than accumulating noise from shared components.

\section{Method}

\begin{figure}[t]
  \centering
  \includegraphics[width=\linewidth]{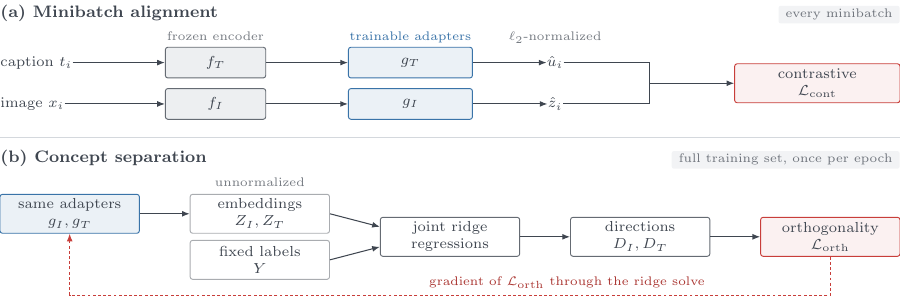}
  \caption{ConceptBasis training. We train two nonlinear projection heads on a frozen vision-language encoder, with contrastive loss on each minibatch and orthogonality loss applied once per epoch.}
  \label{fig:method-overview}
\end{figure}

\subsection{Dictionary and supervision}
\label{sec:dictionary}

\begin{figure}[t]
  \centering
  \includegraphics[width=\linewidth]{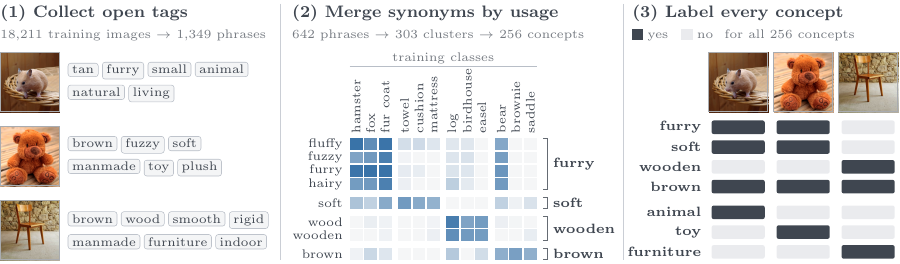}
  \caption{Building the dictionary and its labels. (1) Supply sparse, affirmative visual tags. (2) Cluster interchangeable labels and build a dictionary from the 256 most prevalent clusters. (3) Run an exhaustive annotation pass to build a dense label matrix.}
  \label{fig:dictionary-pipeline}
\end{figure}

A multimodal language model, \href{https://huggingface.co/nvidia/Gemma-4-26B-A4B-NVFP4/tree/a19cfe00be84568a6867111c9a68c9c44fdcffe6}{Gemma 4 26B A4B} in NVFP4 precision \citep{nvidia2026gemma4}, generates grounded captions, discovers candidate concepts, and assigns fixed-dictionary labels (Figure~\ref{fig:dictionary-pipeline}). We construct the dictionary using only training classes. The model first supplies sparse, affirmative visual tags for training images. We cluster tags by their image--text similarity patterns across the training classes and retain the 256 most prevalent clusters as the concept vocabulary. The dictionary is then frozen. A separate exhaustive annotation pass judges every canonical concept for every image in the training, development, and test partitions, as well as the class representatives used for compositional retrieval. \appendixref{apd:dictionary} describes and motivates each step.

\subsection{Estimating partial concept effects}

Let $Z_I,Z_T \in \mathbb{R}^{n\times p}$ be the centered adapted image and paired-caption embeddings, whose rows are the centered $z_i$ and $u_i$, respectively, and let $Y \in \mathbb{R}^{n\times K}$ be the centered concept-label matrix, whose rows are the centered $y_i$. For each modality $m\in\{I,T\}$, we estimate the partial effects of all concepts simultaneously by regressing its embeddings on all observed concept labels with ridge regularization:
\begin{equation}
  B_m=\left(\frac{Y^\top Y}{n}+\alpha \mathbf{I}_K\right)^{-1}
  \frac{Y^\top Z_m}{n}\in\mathbb{R}^{K\times p},
  \qquad m\in\{I,T\},
  \label{eq:ridge}
\end{equation}
where $\alpha>0$ is the ridge penalty and $\mathbf{I}_K$ is the $K\times K$ identity matrix. The $k$th row $b_{m,k}$ of $B_m$ estimates the embedding difference associated with concept $k$ in modality $m$ after controlling for the other observed concepts. Its normalization $d_{m,k}=b_{m,k}/\lVert b_{m,k}\rVert_2$ is the concept direction, and $D_m\in\mathbb{R}^{K\times p}$ collects these directions as rows. Ridge regularization stabilizes directions when labels are correlated or rare. Unlike a group-mean-derived direction, this estimator accounts for the other observed concepts rather than attributing every co-occurring difference to the target concept. The direction-based evaluations below use the image directions, that is, $D=D_I$ in \eqref{eq:coefficient} and \eqref{eq:overlap}.

\subsection{Geometry-aware adapter training}

The image and text adapters are two-layer MLPs trained with the usual symmetric image--caption contrastive loss $\mathcal{L}_{\mathrm{cont}}$. This loss is computed on the $\ell_2$-normalized embeddings $\hat z_i=z_i/\lVert z_i\rVert_2$ and $\hat u_i=u_i/\lVert u_i\rVert_2$, whereas the regression in \eqref{eq:ridge} uses the unnormalized $z_i$ and $u_i$. ConceptBasis applies the orthogonality penalty to both modalities:
\begin{equation}
  \mathcal{L}_{\mathrm{orth}}=
  \sum_{m\in\{I,T\}}
  \frac{1}{K(K-1)}
  \sum_{j\neq k}(d_{m,j}^\top d_{m,k})^2.
  \label{eq:orthogonality}
\end{equation}
Training alternates ordinary minibatch contrastive updates with one full-training-set orthogonality update \eqref{eq:orthogonality} per epoch (Figure~\ref{fig:method-overview}). Contrastive training keeps the adapted image and caption spaces aligned.

\section{Experiments and Results}

\subsection{Setup}
\label{sec:setup}

We use the THINGS dataset \citep{hebart2019things}, with 1,298 object classes for training, 278 disjoint classes for development, and 278 disjoint classes for testing. A frozen SigLIP 2 Giant encoder \citep{tschannen2025siglip2} produces 1,536-dimensional image and caption embeddings. We train a separate nonlinear projection head for each modality:
\begin{equation}
  \mathbb{R}^{1536}
  \xrightarrow{\mathrm{Linear}}
  \mathbb{R}^{1024}
  \xrightarrow{\mathrm{GELU}}
  \mathbb{R}^{1024}
  \xrightarrow{\mathrm{Linear}}
  \mathbb{R}^{256}.
  \label{eq:adapter}
\end{equation}
The image and text heads have the same architecture \eqref{eq:adapter} but separate parameters. \appendixref{apd:implementation} lists all training and evaluation settings.

\subsection{Evaluation protocol and compared methods}

All adapter conditions use the same architecture, data, and minibatch contrastive schedule across five matched random seeds. \emph{Contrastive} trains the adapters only for image--caption alignment and estimates group-mean directions afterward. \emph{Group-mean orthogonality} adds an orthogonality update on those group-mean directions; it is ConceptBasis without the joint-regression estimator. \emph{ConceptBasis} instead estimates all partial-effect directions jointly and orthogonalizes them during training. A further \emph{direct-composition} control uses the same joint-ridge estimator as ConceptBasis but replaces the orthogonality update with sampled compositional retrieval supervision. Finally, SpLiCE is a training-free sparse decomposition of frozen OpenCLIP embeddings. We compare against SpLiCE only for named-coordinate readout because it does not train the shared dense space evaluated by the geometric and compositional tests.

RMS overlap is computed from the train-estimated directions. Ten-attribute retrieval uses one representative from each of the 278 held-out test classes, while ordinary retrieval uses all 3,982 test image--caption pairs. Bidirectional image--text R@5 is the mean of image-to-caption and caption-to-image R@5. Unless stated otherwise, values are mean $\pm$ sample standard deviation over seeds.

\subsection{Main results}

Table~\ref{tab:main-results} compares the three matched training conditions.

\begin{table}[h!]
  \centering
  \caption{Main test results.}
  \label{tab:main-results}
  \small
  \begin{tabular}{@{}p{0.34\textwidth} c c c@{}}
    \toprule
    Model / direction estimator &
    \shortstack{RMS\\overlap $\downarrow$} &
    \shortstack{Compositional R@5\\10 attributes $\uparrow$} &
    \shortstack{Bidirectional\\image--text R@5 $\uparrow$} \\
    \midrule
    Contrastive / group mean
      & $.171 \pm .002$ & $.572 \pm .012$ & $.968 \pm .001$ \\
    Group-mean orthogonality / group mean
      & $.124 \pm .001$ & $.620 \pm .006$ & $.968 \pm .001$ \\
    ConceptBasis / joint-ridge partial effects
      & $\mathbf{.057 \pm .002}$ & $\mathbf{.728 \pm .005}$ & $.967 \pm .002$ \\
    \bottomrule
  \end{tabular}
\end{table}

Relative to orthogonality training on group-mean-derived directions, ConceptBasis reduces RMS overlap from .124 to .057 and raises ten-attribute Recall@5 from .620 to .728, while bidirectional image--text retrieval remains within the variation across seeds.

\subsection{Orthogonality-training ablation}

The matched comparison changes both the orthogonality training and the direction estimator. To isolate the effect of training, we estimate partial-effect directions from every adapter using the same joint ridge regression for direction estimation (Table~\ref{tab:orthogonality-ablation}).

\begin{table}[h!]
  \centering
  \caption{Orthogonality-training ablation using partial-effect directions at evaluation. Whitening replaces the directions with the closest exactly orthogonal set.}
  \label{tab:orthogonality-ablation}
  \small
  \begin{tabular}{@{}p{0.48\textwidth} c c@{}}
    \toprule
    Directions orthogonalized during training &
    \shortstack{RMS\\overlap $\downarrow$} &
    \shortstack{Compositional R@5\\10 attributes $\uparrow$} \\
    \midrule
    None (contrastive only) & $.111 \pm .001$ & $.575 \pm .004$ \\
    None, then whitened after training & $0$ & $.657 \pm .004$ \\
    Group-mean directions & $.109 \pm .001$ & $.518 \pm .004$ \\
    Partial-effect directions (ConceptBasis)
      & $.057 \pm .002$ & $\mathbf{.728 \pm .005}$ \\
    \bottomrule
  \end{tabular}
\end{table}

On the contrastive-only adapter, replacing group means with partial effects reduces overlap from .171 to .111 while leaving compositional retrieval nearly unchanged (.572 versus .575). Controlling for co-occurrence therefore removes substantial statistical overlap but does not by itself produce the full additive gain. With the partial-effect estimator held fixed, group-mean orthogonality offers no improvement, whereas ConceptBasis reaches .057 overlap and .728 Recall@5. This isolates the additional contribution of training the adapter to separate the partial effects themselves. Whitening the contrastive adapter's directions after training makes them exactly orthogonal without changing the embeddings, yet recovers only about half of ConceptBasis's gain (.657 Recall@5), suggesting that orthogonal readout axes cannot fully cancel overlap that remains in the embeddings.

\subsection{Comparison with other methods}

We compare the complete methods along three separate properties. Evaluation classes remain unseen during training.

\subsubsection{Geometric cross-talk}

Directions should be usable individually without moving along other named coordinates through a shared component. RMS overlap measures this geometric cross-talk directly (Table~\ref{tab:cross-talk}).

\begin{table}[h!]
  \centering
  \caption{Geometric cross-talk.}
  \label{tab:cross-talk}
  \small
  \begin{tabular}{@{}p{0.31\textwidth} p{0.39\textwidth} c@{}}
    \toprule
    Method & Direction estimator & RMS overlap $\downarrow$ \\
    \midrule
    Group means (contrastive-only adapter) & Group means & $.171 \pm .002$ \\
    Direct composition & Joint-ridge partial effects & $.104 \pm .001$ \\
    ConceptBasis & Joint-ridge partial effects & $\mathbf{.057 \pm .002}$ \\
    \bottomrule
  \end{tabular}
\end{table}

Direct composition reduces overlap incidentally, but leaves almost twice as much as ConceptBasis. Its retrieval objective can therefore succeed without producing the same separated geometry.

\subsubsection{Additive composition}

Directions should also combine without repeatedly counting shared components. We test this with ten-attribute compositional retrieval on the 278 class-disjoint test images (Table~\ref{tab:additive-composition}).

\begin{table}[h!]
  \centering
  \caption{Additive composition and ordinary retrieval.}
  \label{tab:additive-composition}
  \small
  \begin{tabular}{@{}p{0.38\textwidth} c c@{}}
    \toprule
    Method & Compositional R@5 $\uparrow$ & Bidirectional image--text R@5 $\uparrow$ \\
    \midrule
    Group means (contrastive-only adapter) & $.572 \pm .012$ & $.968 \pm .001$ \\
    Direct composition & $\mathbf{.822 \pm .002}$ & $.967 \pm .001$ \\
    ConceptBasis & $.728 \pm .005$ & $.967 \pm .002$ \\
    \bottomrule
  \end{tabular}
\end{table}

While direct composability training predictably performs the best on the task it explicitly optimizes, ConceptBasis nevertheless captures most of this improvement without compositional supervision, while producing substantially less direction overlap. Bidirectional image--text retrieval is nearly unchanged across the three conditions.

\subsubsection{Named-coordinate readout}

Each coordinate should predict the presence of its named concept. We measure macro AUROC on all 3,982 test images, restricting every method to the 174 concept names shared with SpLiCE's native LAION-10K vocabulary.

Group means perform best on named-coordinate readout ($.907\pm.001$), as expected from an estimator that can use correlated concepts as predictive signal. ConceptBasis retains strong readout ($.868\pm.001$) while producing much lower direction overlap. SpLiCE reaches $.536$ with its native OpenCLIP setup and optimizes reconstruction rather than individual coefficient meanings, allowing correlated or more specific atoms to substitute for the exact named atom. Table~\ref{tab:named-coordinate} reports the complete comparison.

At sparsity penalty $\lambda=.25$, SpLiCE activates 13.5 atoms per image on average. Nevertheless, the exact-name atom is zero for 92.7\% of positive images on average across concepts, and 41 of 174 concepts never activate on any positive test image. Table~\ref{tab:splice-substitution} gives development examples in which nearby atoms reconstruct the embedding while the intended named coefficient remains zero.

\section{Limitations}

Image labels are synthetic data produced by a multimodal language model and may inherit its annotation errors.

Estimated partial effects control only for concepts present in the dictionary, so the adapter is not penalized for overlap with omitted concepts. Therefore, ridge regression does not isolate a perfect causal effect.

\section{Conclusion}

Concept directions derived from linear probes or differences between group means are widely used for readout, steering, and debiasing. However, existing techniques produce directions for semantically independent concepts containing substantial components of one another, resulting in cross-talk during editing and repeated counting of shared components during additive composition. We distinguish this goal from alternatives such as SpLiCE and direct composability training. ConceptBasis estimates partial concept effects with ridge regression and trains lightweight adapters to reduce their remaining overlap, substantially reducing cross-talk and improving additive retrieval while preserving bidirectional image--text retrieval and named-coordinate readout. In conclusion, we show that the linear representation hypothesis can be extended to create an additive, separable coordinate system of semantic concepts.

\FloatBarrier
\bibliography{references}

\appendix
\section{Dictionary construction and labels}
\label{apd:dictionary}

ConceptBasis needs two things from the data: a fixed vocabulary of visual concepts, and a yes-or-no label for every concept on every image. We build both in three steps, illustrated in Figure~\ref{fig:dictionary-pipeline}. The Gemma model of Section~\ref{sec:dictionary} proposes candidate properties for the training images; we merge phrases that name the same property and keep the 256 most prevalent; and the same model then labels every image against this fixed list. The full prompts and the construction script are in the released code.

\paragraph{Proposing candidate properties.} Rather than choosing concepts by hand, we let the model propose them. For each training image, it receives the image and its THINGS class name, which identifies the object to describe, and lists 5--50 properties of one or two words, covering appearance (color, material, finish, texture, shape) and kind or function (for example, animal, food, tool, or wearable). The prompt asks for properties that are neither specific to one kind of object nor true of nearly every object: a concept that is almost always absent or almost always present says little about any particular image and carries little information into the embedding. It also asks only for properties that are present, excluding object-specific parts, the object's name, non-visual facts, and negations, so that every concept denotes a trait the object has rather than one it lacks. Sampling uses temperature 0.3. Only the training split takes part: splits are class-disjoint, with every image of a class in the same split (Section~\ref{sec:setup}), so development and test classes never influence the vocabulary. The 18,211 training images receive 8.2 tags on average (range 5--22), giving 1,349 distinct phrases. Phrases tagged on fewer than 3 images or in fewer than 3 classes would be absent almost everywhere, so we drop them, together with explicit lexical negations of another candidate (here only \texttt{unpolished}), leaving 642 phrases.

\paragraph{Merging synonyms.} Open tags often name one property in several words, such as \texttt{fuzzy}, \texttt{furry}, and \texttt{hairy}. Kept apart, these would become separate concepts whose directions the orthogonality objective pushes apart, even though they describe the same property. We therefore merge phrases by how they are used rather than how they are spelled. The frozen SigLIP 2 text encoder embeds each phrase $w$ as ``an object that is $w$'', and a phrase's usage profile is its mean image--phrase similarity within each of the 1,298 training classes, standardized across classes. Phrases with matching profiles describe the same classes, and we treat them as naming the same property. Before comparing profiles, we remove the first principal component of the class-by-phrase matrix: some classes' images score uniformly higher or lower against every phrase, and this shared offset would otherwise make nearly all profiles look alike. We then cluster the phrases by complete-linkage agglomerative clustering on one minus the Pearson correlation between profiles, cut at $0.5$. Complete linkage requires every pair of phrases in a cluster to correlate at least $0.5$, so clusters cannot grow by chaining through intermediate phrases: \texttt{soft} correlates with \texttt{fluffy} but not with \texttt{furry}, and remains a separate concept. This yields 303 clusters.

\paragraph{Choosing the dictionary.} At most $p$ directions can be mutually orthogonal, so we keep $K=256$ concepts, as many as the $p=256$ embedding dimensions. To decide which, we measure a phrase's class-balanced prevalence as the fraction of each class's images tagged with it, averaged over classes, so that classes with many photographs do not dominate; a cluster's prevalence counts an image tagged with any of its members. Each cluster is named after its most prevalent member, and clusters are ranked by prevalence. Walking down this ranking, we keep each cluster unless its profile correlates above $0.75$ with a concept already kept, which removes near-duplicates that survived clustering, such as \texttt{ornamental} after \texttt{decorative} (4 in total), and we stop at 256 concepts. Of these, 85 are single phrases and the largest merges 9.

\paragraph{Labeling every concept.} Open tags are too sparse to use as labels. An image's tags omit most of the properties that apply to it, and the joint regression~\eqref{eq:ridge} would read every omission as absence. A second pass therefore asks about every concept for every image, in all splits and for the CC0 class representatives used in compositional retrieval. The model receives the image, its class name, and the full list of 256 concepts, and answers yes or no for each one. Because photographs often include other objects and backgrounds, it judges only the named object, not the background, other objects, or lighting. Decoding is deterministic (temperature 0, reasoning disabled). Training images receive 22.5 positive labels on average (range 3--48), nearly three times their 8.2 open tags.

\section{Implementation details}
\label{apd:implementation}

Table~\ref{tab:implementation} lists the training and evaluation settings for every adapter condition. All adapter results were produced with commit \href{https://github.com/markjouh/conceptbasis/tree/786074beed8c5d23681731fbddeecc54195f0188}{\texttt{786074b}} of the released code, whose \texttt{reproduce.sh} regenerates them from the released dictionary, fixed labels, and THINGS images. The SpLiCE results have their own script in the same repository.

\begin{table}[h!]
  \centering
  \caption{Training and evaluation settings.}
  \label{tab:implementation}
  \small
  \begin{tabular}{@{}p{0.28\textwidth} p{0.68\textwidth}@{}}
    \toprule
    Setting & Value \\
    \midrule
    \multicolumn{2}{@{}l}{\emph{Shared by all adapter conditions}} \\
    Training data & 18,211 images and their grounded captions from the 1,298 training classes \\
    Encoder & Frozen SigLIP 2 Giant (\href{https://huggingface.co/timm/ViT-gopt-16-SigLIP2-384/tree/ad3410bee2c3373be5ed01e7c4e7fcd2bf95a183}{\texttt{timm/ViT-gopt-16-SigLIP2-384}}, revision \texttt{ad3410b}); 1,536-dimensional image and caption embeddings \\
    Adapters & Separate image and text MLPs \eqref{eq:adapter}; $p=256$ \\
    Contrastive loss & Symmetric InfoNCE (softmax cross-entropy over in-batch pairs) on $\hat z_i$ and $\hat u_i$; learnable inverse temperature initialized to $1/0.07$ and capped at 100 \\
    Optimizer & AdamW; learning rate $10^{-3}$; weight decay $10^{-4}$; cosine annealing over 30 epochs \\
    Schedule & 30 epochs of minibatch contrastive updates on 1,024 image--caption pairs; after each epoch, one full-training-set concept update (below) \\
    Seeds & 0--4, matched across conditions \\
    \midrule
    \multicolumn{2}{@{}l}{\emph{Per-epoch concept update}} \\
    Contrastive & None \\
    Group-mean orthogonality & One optimizer step on $\lambda_{\mathrm{orth}}\mathcal{L}_{\mathrm{orth}}$ \eqref{eq:orthogonality}, with each direction replaced by the normalized difference between the mean embeddings of images with and without the concept, in each modality; $\lambda_{\mathrm{orth}}=8$ \\
    ConceptBasis & One optimizer step on $\lambda_{\mathrm{orth}}\mathcal{L}_{\mathrm{orth}}$ \eqref{eq:orthogonality}, with gradients through the ridge solution \eqref{eq:ridge} into both adapters; $\lambda_{\mathrm{orth}}=1024$ \\
    Direct composition & One optimizer step on a retrieval loss with weight 1: for each of 1,024 sampled training images, a random subset of 1--14 of its present concepts is summed, with directions standardized by their training projection scale, and must retrieve its source image among the sampled images \\
    Ridge penalty & $\alpha=0.01$, applied after dividing the centered $Z_m$ by a single gradient-free scalar so that their root-mean-square entry is 1; this rescaling does not change the directions \\
    \midrule
    \multicolumn{2}{@{}l}{\emph{Evaluation}} \\
    Directions & Re-estimated from the final adapters on the full training set with the estimator named in each table; $\mu_k$ and $\sigma_k$ in \eqref{eq:coefficient} from training-set projections \\
    Compositional retrieval & Gallery of the 278 CC0 test-class representatives; for each representative, 24 fixed-seed random orderings of its present concepts define nested queries $S$; Recall@5 over queries with $\lvert S\rvert=10$ \\
    Ordinary retrieval & All 3,982 test image--caption pairs; mean of image-to-caption and caption-to-image Recall@5 \\
    Coordinate readout & Macro AUROC of $c_k(x)$ over the 3,982 test images, for concepts with both positive and negative test images \\
    \bottomrule
  \end{tabular}
\end{table}

\clearpage
\section{Additional SpLiCE results}

\begin{table}[h!]
  \centering
  \caption{Named-coordinate readout on the 174 concepts shared with SpLiCE.}
  \label{tab:named-coordinate}
  \small
  \begin{tabular}{@{}p{0.36\textwidth} p{0.28\textwidth} @{\hspace{1.5em}} r@{}}
    \toprule
    Method & Coefficient & Coordinate AUROC $\uparrow$ \\
    \midrule
    Group means (contrastive-only adapter) & Standardized projection & $\mathbf{.907 \pm .001}$ \\
    Direct composition & Standardized projection & $.900 \pm .002$ \\
    SpLiCE (native LAION-10K, $\lambda=.25$) & Sparse atom weight & $.536$ \\
    ConceptBasis & Standardized projection & $.868 \pm .001$ \\
    \bottomrule
  \end{tabular}
\end{table}

\begin{table}[h!]
  \centering
  \caption{Development examples of coefficient substitution in SpLiCE.}
  \label{tab:splice-substitution}
  \small
  \begin{tabular}{@{}p{0.14\textwidth} p{0.24\textwidth} p{0.50\textwidth}@{}}
    \toprule
    \shortstack[l]{Positive\\image} & \shortstack[l]{Zero exact-name\\coefficient} & Largest nonzero atoms include \\
    \midrule
    \texttt{tortilla} & \texttt{edible} & \texttt{tortilla}, \texttt{recipe}, \texttt{pancakes} \\
    \texttt{stereo} & \texttt{electronic} & \texttt{stereo}, \texttt{radio}, \texttt{cassette} \\
    \texttt{strawberry} & \texttt{fruit} & \texttt{strawberry}, \texttt{nutritional}, \texttt{produce} \\
    \texttt{tiger} & \texttt{animal} & \texttt{tiger}, \texttt{snowy}, \texttt{threatened}, \texttt{wildlife} \\
    \bottomrule
  \end{tabular}
\end{table}

\end{document}